\documentclass[sn-mathphys]{sn-jnl}% Math and Physical Sciences Reference Style
\jyear{2021}%
\usepackage{natbib}
\usepackage{arabtex}
\usepackage{utf8}
\usepackage{graphicx} % include this line in the preamble
\usepackage[utf8]{inputenc}
\theoremstyle{thmstyleone}%
\theoremstyle{thmstyletwo}%

\theoremstyle{thmstylethree}%

\usepackage[utf8]{inputenc}
\UseRawInputEncoding

\begin{document}

\title[Arabic Hate Speech Identification and Masking]{Arabic Hate Speech Identification and Masking in Social Media using Deep Learning Models and Pre-trained Models Fine-tuning}


\author*[1]{\fnm{Salam Thabet} \sur{Doghmash}}\email{salamthabet@students.iugaza.edu.ps}

\author[1]{\fnm{Motaz} \sur{Saad}}\email{msaad@iugaza.edu.ps}
\equalcont{These authors contributed equally to this work.}

\affil*[1]{\orgdiv{Department of Data Science}, \orgname{Faculty of Information Technology, Islamic University of Gaza}, \orgaddress{\street{Jama Abdelnaser St.}, \city{Gaza}, \postcode{108}, \state{Gaza Strip}, \country{Palestine}}}


\abstract{
Hate speech identification in social media has become an increasingly important issue in recent years. In this research, we address two problems: 1) to detect hate speech in Arabic text, 2) to clean a given text from hate speech. The meaning of cleaning here is replacing each bad word with stars based on the number of letters for each word. Regarding the first problem, we conduct several experiments using deep learning models and transformers to determine the best model in terms of the F1 score. Regarding second problem, we consider it as a machine translation task, where the input is a sentence containing dirty text and the output is the same sentence with masking the dirty text. The presented methods achieve the best model in hate speech detection with a 92\% Macro F1 score and 95\% accuracy. Regarding the text cleaning experiment, the best result in the hate speech masking model reached 0.3 in BLEU score with 1-gram, which is a good result compared with the state of the art machine translation systems.}




\maketitle

\section{Introduction}\label{sec1}

With the rise of social media platforms, hate speech has become more visible and accessible than ever before. This has led to a growing need for automated methods to identify, monitor and moderate hate speech on social media platforms. Such methods can help to reduce the spread of hate speech, as well as provide valuable insights into the prevalence of hate speech on these platforms. There are many challenges associated with hate speech identification such as the Ambiguity of the meaning because statements may have different meanings in different contexts, the subjectivity of identifying hate speech posts among human themselves as different people may interpret the same statement differently, and the False positives as automated systems may flag innocent posts as hatred speech. 

Hate speech is a crime that has been on the rise in recent years, not just in reality but also on online platforms \cite{fortuna2018survey}. Several factors contribute to this, increasing the number of people who use the internet, social media platforms that have helped it spread dramatically, and migration and growing conflicts around the world, which encourage people to express their opinions online, contributing to the spread of hate speech, thus contributing to the propagation of hate speech as well. Hate speech is defined as any expression that degrades an individual or group in terms of certain factors like a racial religious or national appearance that’s a gender related sexual identity or other identities \cite{Fortuna2017AutomaticDO} \cite{basile-etal-2019-semeval}. The challenge of identifying hate speech and cleaning content has become increasingly important in recent years when hate speech has become the norm.
The work on Arabic offensive language detection is relatively nascent with the high attention focused on English \cite{inproceedings}. Online communities, social media platforms, and technology companies have been investing heavily in ways to cope with offensive language to prevent abusive behavior in social media. In this research, we address two problems: the first one is Arabic hate speech detection by using different neural networks architectures, which applies it to a dataset from shared task SemEval-2020 for Arabic offensive language detection \cite{Zampieri}. 
In the second problem in this research, we address the problem of cleaning offensive/hate speech texts. The idea is to consider the problem of cleaning dirty text as a machine translation problem, instead of providing the text in the source language to the MT system to produce the text in the target language, the input is dirty text that defined as the text that contains a feature of hate speech, sexual harassment, offensive, verbal abuse for disability or disease, gender bias, race bias, national origin, religion, and the output is clean text. To the best of our knowledge, works in literature focus on hate speech detection, and there is no research which addresses the problem of cleaning such texts. In order to accomplish so, we need a parallel corpus. A parallel corpus is a collection of this texts, each of which is translated into one or more other languages than the original \cite{Teubert}. To build such corpus, we mainly need human workers to mask the dirty word from sentences by replacing each word with stars based on the number of letters for it.

The rest paper organized as the following Section \ref{sec:related works} Reviews related works, Section \ref{sec:hate speech detection} Describe hate speech detection, Section \ref{sec: hatespeech masking} Describe hate speech masking and Section \ref{sec: conclusion} presents conclusion and future works

\section{RELATED WORKS}\label{sec:related works}

This section reviews the recent and relevant research in the field of hate speech detection and masking. We review the previous works as follows: First, there are works that construct new datasets for hate speech detection tasks. Second, there are works that focus on generating hate/obscene/offensive terms automatically. Third, there are works that use ML, DL, and transformer learning to detect hate speech. finally, there are systematic/ analytical studies.

\subsection{Construction datasets for hate speech detection tasks}\label{subsec2}

\cite{mulki-etal-2019-l} Gathered dataset from Twitter for identifying hate speech and abusive language in Arabic text, and it is available as a benchmark dataset called L-HSAB. There were 5,846 tweets in this dataset, which were divided into three categories: hate, normal, and abusive. They used ML classification using NB and SVM classifiers in experiments on their dataset. Using the NB classifier, the results were: 90.3\%, 89.0\%, 90.5\%, and 89.6\% in accuracy, recall, precision, and f1-measure.

\cite{Albadi2018AreTO} Investigate the challenge of religious hate speech detection in Arabic Twitter which is the first effort in this field. By creating and publishing a dataset of 6,136 tweets, approximately 1000 for each of the six religious groups labeled as hate or not hate (Muslims, Jews, Christians, Atheists, Sunnis, and Shia). Then they produced and published three lexicons of religious hate phrases that can be used for a variety of purposes, including sampling microposts for religious hate speech. Finally, they looked into three methods for detecting religious hate speech: lexicon-based, n-gram-based, and deep learning-based methods. With 0.79 accuracy and 0.84 AUROC, the GRU-based RNN with pre-trained word embeddings had the best results.

\citep{mubarak-etal-2017-abusive} Provides an automated method for creating and extending a list of obscene terms. First, they collect 175 million tweets in the Arabic language as an initial data set, by searching for some patterns that are usually used in offensive communication, The words that appeared following these patterns were then gathered and personally analyzed to see if they were obscene or not, resulting in a final list that included 415 words after adding hashtags that are used to screen pornographic pages, second, from the same initial set they classified twitter users to two groups: the clean group who authored tweets that did not include a single obscene word, and obscene group who used at least one of the obscene words, the Log Odds Ratio (LOR) was then calculated for each word unigram and bigram that appeared at least ten times. The tweets written by clean tweeps work as a background corpus, whereas the tweets written by obscene tweeps work as a foreground corpus. Additional 3,430-word unigrams and bigrams were formed by the unigrams and bigrams that yielded a LOR equals infinity, which signifies they only appeared in the foreground corpus (obscene) but not in the background corpus (clean). The authors collected other datasets from a popular Arabic news site by capturing user comments that were deleted from the website then they violated the site’s rules.

\subsection{Detection of hate speech using ML, DL, Transformer techniques}\label{subsubsec2}

\cite{Saksesi2018} Proposed using a deep learning method with the Recurrent Neural Network (RNN) to determine whether or not the text contains hate speech. The total twitter dataset that uses it is 1235 records and there are 652 records classified as hate speech and 583 records are not hate speech. They had made the test with several techniques such as Recurrent Neural Network, Data Partition, Epoch, Learning Rate, and Batch Size. The testing results reached 91\% precision, 90\% recall, and 91\% accuracy on average.

\cite{Qiang2021} proposed using a deep learning model namely Convolutional Neural Network for classification, this classifier assigns each tweet to one of the classes of a twitter dataset: hate, offensive, or neither. The accuracy, precision, recall, and F-score of this model have all been used to assess its performance. The final model has a 91\% accuracy, 91\% precision, 90\% recall and an F-measure of 90\%. It should be noted that it is also suggested to further analyze the predictions and errors, to realize more insight on the misclassification.

\cite{alami-etal-2020-lisac} Proposed using AraBERT for the identification of offensive language from Arabic content. Starting from preprocessing tweets by dealing with emojis and replacing them with their meanings in Arabic. Then, in both the fine-tuning and inference steps, they replaced any emojis with the token [MASK]. The AraBERT concept was then applied to tweet tokens. Finally, they pass them into a sigmoid function to determine if a tweet is offensive or not. This approach achieved the best macro F1 score equal to 90.17\% on the Arabic task in OffensEval 2020.

\cite{hassan-etal-2020-alt-semeval} Performed many experiments for offensive language identification in Arabic with SVMs, DNNs, and Multilingual-BERT. The best results were obtained by the aforementioned models using an ensemble approach based on majority vote. With a macro F1 score of 90.16\%, this model came in second in the official rankings.

\cite{wang} Provided a multilingual method using pre-trained language models ERNIE and XLM-R, their technique has two phases, starting with pre-training using large scale multilingual unsupervised texts, which results in a unified pre-training model that learns all language representation at the same time. Then used labelled data to fine-tune the pre-trained model. This technique obtained an F1 macro score of 0.89 on Arabic.

\cite{Safaya} Proposed an approach using Convolutional Neural Networks with a pre-trained BERT model for the offensive language identification task from SemEval 2020 \cite{Zampieri}. They prove that combining BERT with CNN outperforms using BERT alone, and they highlight the necessity of employing pre-trained language models for downstream tasks. This approach acquired a macro averaged F1-Score of 0.897 in Arabic, which ranked fourth among participating teams for the Arabic language in the scope of the OffensEval 2020.

\cite{keleg-etal-2020-asu} Presents different models for offensive language detection. These models are the TF-IDF and logistic regression, CNN using word embeddings from Aravec, Bi-directional LSTM using word embeddings from Aravec, fine-tuning multilingual BERT, and fine-tuning AraBERT. They've also created a list of obscenity words and utilized simple augmentation rules to construct the many variants of each. The AraBERT-based model, which outperformed the cased multilingual BERT model, was their best model. This system ranked fifth in the official rankings, with a macro F1 score of  89.6\%. 

\cite{mnassri2023hate} Develops a multi-task learning (MTL) model to detect parts of hate speech in addition to offensive language using BERT. This model has been built with the objective of enhancing performance by enabling features across tasks such as hate speech detection, offensive language detection, and emotion recognition to be shared. The model is built on BERT sharing and allows each task to benefit from common features. They used contextual representations of text to enhance hate speech and offensive language detection accuracy. Using this model in multiple experiments with various datasets, they have proven that the multi-task model could achieve better results than other models in identifying hate speech and offensive language.

\cite{Detecting2019} Offers many various neural networks models for detecting offensive language on Arabic social media. These models are CNN, bidirectional LSTM, and merged CNN-LSTM. These models are evaluated on an Arabic YouTube comments dataset that include 15,050 comments extracted from famous and contentious YouTube videos with Arab celebrities. They employ Arabic word embeddings to represent the comments and train this dataset through a set of pre processes. The combined CNN-LSTM network has the best recall of 83.46\% while the CNN has the best accuracy of 87.84 and precision of 86.10.

\cite{Fatemah} Proposed applying transfer learning to several Arabic offensive language datasets separately and testing it with other datasets separately, as well as investigating the impacts of concatenating all datasets to be utilized for fine-tuning and training the BERT model. These datasets involve Aljazeera.net Deleted Comments \cite{mubarak-etal-2017-abusive}, YouTube dataset \cite{ALAKROT2018174}, Levantine Twitter Dataset for Hate Speech and Abusive Language (L-HSAB) \cite{mulki-etal-2019-l}, and OSACT offensive and not offensive classification samples \cite{hassan-etal-2020-alt-semeval}. They totally depend on binary classes; offensive or non-offensive, and they change various types of offensive languages like abusive or hate to the offensive class. The highest recorded scores are shown for the OSACT dataset when used in training and testing. Their findings show that Arabic monolingual BERT models outperform BERT multilingual models, and that transfer learning across datasets from multiple sources and topics, such as YouTube comments from musicians channels and Aljazeera News comments from political articles, performs poorly. When comparing individual datasets, combining from multiple datasets at the same time has no effect on performance; however, it affects the performance of the highly dialectic dataset, L-HSAB, by 3\% in macro F1 score.

\cite{Alshalan} Provides dataset collected from Twitter to detect hate speech, this dataset contains 9316 tweets labeled as hateful, abusive, and normal. Then they evaluated different Deep neural network models based on CNN and RNN, these models are CNN, GRU, CNN + GRU, and BERT. The results appear that CNN outperformed other models, with an F1-score of 0.79 and an AUROC of 0.89, whereas BERT failed to increase performance, which might be due to the fact that BERT was trained on Wikipedia, which is a different kind of dataset.

\cite{icpram20} Collected dataset from Twitter that included hate expressions on a variety of topics in the Arabic language. This data was gathered using various terms such as racism, sport, and Islam, and then categorised as Hate or Normal. The authors proposed using a deep learning approach for the automatic detection of cyberhate speech. This approach combines a convolutional neural network (CNN) and a long short-term memory (LSTM) network with the Word2Vec and AraVec word embedding techniques to extract a set of words features that can take the hidden relations of words in the dataset. The proposed method performed well in identifying tweets as Hate or Normal, with the best one scoring 66.564\%, 79.768\%, 68.965\%, and 71.688\% for the accuracy, recall, precision, and F1 measure.

\section{HATE SPEECH DETECTION}\label{sec:hate speech detection}

This section describes the methodology that we use for hate speech detection. The methodology steps including dataset description, preprocessing techniques, text features, and the models that we use for hate speech detection. Finally, the section presents experimental setups, evaluation metrics, results and discussion.

\begin{figure}[h]
    
    \centering
    \includegraphics[scale=1.2]{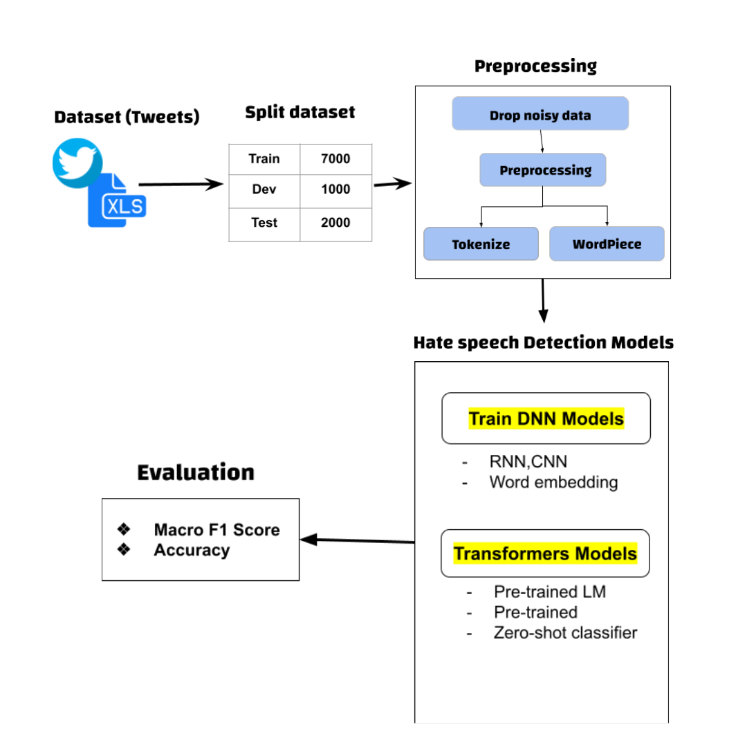}
    \caption{The Brief of Hate Speech Detection Methodology}
    \label{fig:methodology}
\end{figure}

Figure \ref{fig:methodology} presents a brief overview of the methodology phases including data acquisition, data preprocessing, training models, and evaluation metrics.

\subsection{Dataset}
We use the dataset that is published in the shared task SemEval-2020 \cite{Zampieri} for Arabic offensive language detection. This dataset was collected from Twitter and contains 10000 tweets labeled either for offensive or not offensive. The dataset is partitioned into 7000 tweets for training, 1000 tweets for development, and 2000 for testing, just like the SemEval competition. Table 1 shows the details of the dataset parts. As shown in the Table 1, class distribution is imbalanced, i.e there are only 1991 offensive tweets vs 8000 not offensive.

\begin{table}[h]
\begin{center}
\begin{minipage}{174pt}
\caption{The dataset distribution}\label{tab1}%
\begin{tabular}{@{}llll@{}}
\toprule
 & Training set & Development set & Test set\\
\midrule
Offensive   & 1410   & 179  & 402  \\
Not offensive    & 5590   & 821  & 1598  \\
Total    & 7000   & 1000 & 2000  \\
\botrule
\end{tabular}
\end{minipage}
\end{center}
\end{table}

\subsection{Data Preprocessing}

This step is important for cleaning data from unnecessary content and transforming it into a consistent format that can be simply processed and analyzed. In our work, we used classical text preprocessing steps as following:

1.	Letter normalization: which means the process of transforming letters that appear in different forms into a single form. The normalization step includes: replace
(\RL{أ ، إ ، ا، آ }) with (\RL{ة}), (\RL{ا}) with (\RL{ه}), and (\RL{ى}) with (\RL{ي}), the purpose of this step is to reduce the orthographic differences that can be seen in tweets.

2.	Remove punctuations and diacritics: We exclude question marks and exclamation marks from this step.

3.	Remove repeating characters.

4.	Remove all words that contain non-Arabic characters.

\subsection{Tokenize and AraBert WordPiece}

The WordPiece representation was created to automatically learn word-by-word from large amounts of data and not generate OOV. This technique for handling OOV is used in BERT. OOV is ignored in word2vec and GloVe, but the letter ngram representation of a word in FastText corrects OOV. WordPiece tokenization splits a word into different tokens. The most important words are retained and the other words are subdivided \cite{Alyafeai}. In the transformer models, the text is passed to a transformer model divided by the text with WordPiece.

Text features are represented as word embeddings through Word2Vec. We use word2vec model to load Twt-CBOW \cite{Mohammad}.

\subsection{Hate speech detection models}
We investigate different neural networks architectures for detection, these models include Recurrent Neural Network (RNN), Convolutional Neural Network (CNN), and Transformers.

In our hate speech detection experiments we applied two strategies, the first strategy is building and training deep learning models DL from scratch, and the second one uses pre-trained and transformer models.

\subsubsection{Building RNN Model}

\begin{figure}[h]
    \centering
    \includegraphics[scale=0.34]{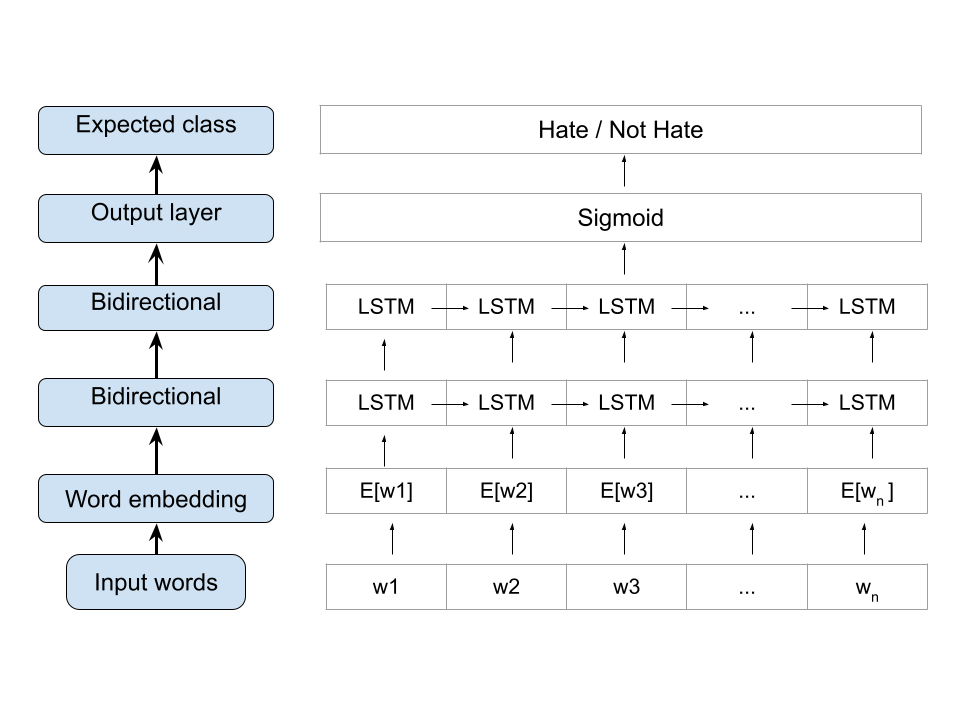}
    \caption{RNN architecture: Left represent RNN layers and right represent layers details}
    \label{fig:RNN}
\end{figure}

We selected the RNN architecture for this challenge because sequential information is important in detecting hate speech sentences. We applied a model from TensorFlow that uses RNN for text classification. This model architecture contains four layers: an input layer (embedding layer), a two Bidirectional LSTM layer, and finally the output layer, as represented in Figure \ref{fig:RNN}, the input tweets are fed into the embedding layer, which maps tweets tokens into a 300-dimensional real-valued vector. The embedding layer produces an output matrix, this output matrix is received by two parallel Bidirectional LSTM layers with 128, 64 units sequentially, the shape that is produced is passed to a dropout layer with a rate of 0.5 is used to reduce the overfitting problem. The final layer is a sigmoid layer that produces the final predictions. In this experiment we use the following parameters: batch size 1024, RNN sequence length 25, number epochs 30, Learning Rate is 1e-3, and Adam optimizer.

\newpage
\subsubsection{Building CNN Model}
\begin{figure}[h]
    \centering
    \includegraphics[scale=0.34]{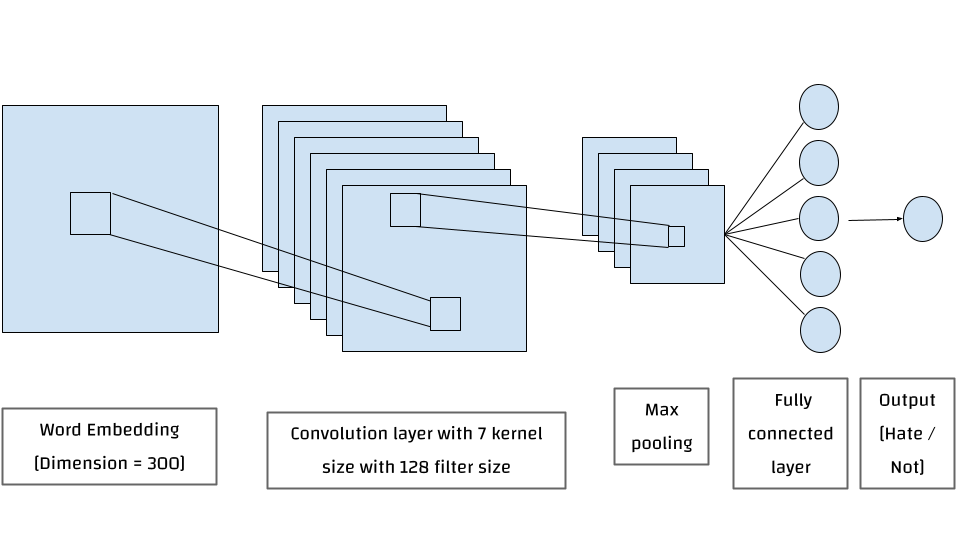}
    \caption{CNN architecture for text classification. CNN layers include word embedding as input layer, Conv1D, max pooling, fully connected, and output layer}
    \label{fig:CNN}
\end{figure}

Figure \ref{fig:CNN} illustrates our CNN architecture, which includes five layers: an input layer (embedding layer), a convolution layer, a pooling layer, a hidden dense layer, and finally the output layer. Here Similar to RNN architecture we refer to that all tweets were mapped into 300-dimensional real-valued vectors by the embedding layer, the embedding layer then passes an input feature matrix to a dropout layer with a rate of 0.5. The dropout layer's primary objective is to help prevent overfitting issues. Then, the output is received by the convolution layer that has 128 filters with the same kernel sizes: 7 and a rectified linear unit (ReLU) function for activation. After that, these convolution features are fed as input to a max-pooling layer (global) for downsampling, then concatenated and passed as input to a fully connected dense layer containing 128 neurons followed by a dropout layer with a rate of 0.5. The output is then fed to the output layer with sigmoid activation to produce the final predictions.

\subsubsection{Building CNN-RNN Model}
\begin{figure}[h]
\centering
\includegraphics[scale=0.5]{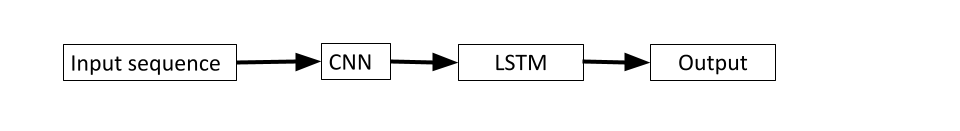}
\caption{A combination between CNN with RNN: CNN block represents all CNN layers in Figure \ref{fig:CNN} except for the output}\label{fig:cnnrnn}
\end{figure}

In this model, we make a combination of both CNN and RNN as shown in Figure \ref{fig:cnnrnn}. The CNN and RNN combination architecture contain six layers: an input layer (embedding layer), a convolution layer with 128 filters and a kernel size of 7, a max-pooling layer, a Bidirectional LSTM layer, another LSTM layer, and finally the output layer. The embedding layer starts by mapping the tweets into a 300-dimensional vector space, producing a tweets matrix. This matrix is then passed to a dropout layer with a rate of 0.2 to avoid the overfitting problem. Then, the output of the dropout layer is fed into the convolution layer, which has 128 filters with kernel sizes of 7. The rectified linear unit (ReLO) function is used for activation. then passed into a max-pooling layer with a pool size of 2 and a dropout layer with a rate of 0.2. This produces vector output, which can be considered as extracted features. These extracted features are then passed to the RNN (Bidirectional LSTM) layer with 128 units, followed by the LSTM layer also with 128 units, then followed by a dropout layer with a rate of 0.2. Finally, the output of the dropout layer is then fed into the output layer with sigmoid activation to produce the final predictions.

\subsubsection{Using Transformer pre-trained models}

In this section, we describe the transformer models that are use in our experiment. Starting from Arabic Pre-trained language models QARiB \cite{QARiB}, \cite{MARBERT}, and Multi-dialect-bert-base-arabic. Then, pretrained hate speech models xlm-r-large-arabic-toxic and dehatebert-mono-arabic, finally, that Zero-shot classifier models Xlm-roberta-large-xnli, XLM-RoBERTa-large-XNLI-ANLI, and Roberta-large-mnli.

All transformer models that we use are available through the HuggingFace   Transformers library.

\textbf{Arabic Pre-trained language models}

In our experiments use three Arabic pre-train language models QARiB, MARBERT, and Multi dialect Arabic BERT using hugging-face API; Table 4 shows the model names used and their details. We use the same Arabert implementation that available in the Github repository  to load models, and the parameters that we used are highlighted in Table 2. Then, we train and evaluate these models on the dataset that was describe before.

\begin{table}[h]
\begin{center}
\begin{minipage}{174pt}
\caption{Parameters value that we used in pretrained models
}\label{tab1}%
\begin{tabular}{@{}llll@{}}
\toprule
 Parameter & Value \\
\midrule
Epsilon (Adam optimizer)   & 1e-8   \\
Learning Rate    & 5e-5    \\
Batch Size     & 16  \\
\#Epochs    & 8  \\

\botrule
\end{tabular}
\end{minipage}
\end{center}
\end{table}

\textbf{Hate Speech Pretrained models}

Hate Speech pre-trained models designed to classify hate speech and detect toxic content in the Arabic language. The first model we use it in our experements is dehatebert-mono-arabic , and the other model is xlm-r-large-arabic-toxic. 

The pre-trained models are used in two ways, the first is training from scratch, and the second way is transfer learning which means freeze some layers, and do additional training to the rest of layers to tune the model for the new domain/task/data. The parameters that we used are highlighted in Table 3.

\begin{table}[h]
\begin{center}
\begin{minipage}{174pt}
\caption{Parameters value that we used in fine tuning hate speech pretrained models
}\label{tab1}%
\begin{tabular}{@{}llll@{}}
\toprule
 Parameter & Value \\
\midrule
Learning Rate    & 3e-5    \\
Batch Size     & 7  \\
\# Epochs    & 3  \\

\botrule
\end{tabular}
\end{minipage}
\end{center}
\end{table}

\textbf{Zero shot classifier models}

The zero-shot text classification model makes a big difference in technology because it can classify any text into any category without prior data. To perform and inference(predict) with the zero-shot classifier, we need to pass the text and the candidate labels. We try many candidate labels and see which labels produce the best results. The Zero-Shot classifiers are xlm-roberta-large-xnli, roberta-large-mnli, and xlm-roberta-large-xnli-anli.

\subsection{Evaluation metrics}

Evaluation metrics considered in this study (Macro F1-score) as similar to those used in the literature for the Shared Task, also we recorded precision, recall, f1 score, and accuracy for each model.

\subsubsection{Experiment and Results}

We perform a set of experiments to evaluate the above mentioned models for Arabic hate speech detection tasks using deep learning models. In all experiments, we consider a binary classification task in which tweets were classified to one of these classes (offensive or not-offensive, hate or not-hate, or toxic or not toxic). 

In the following subsections describes the results of experiments of all models described in the previous section.

\vspace{5mm} %5mm vertical space

\paragraph{Results of DL models}

In DL models experiments, the results are very close to each other, and we notice models that use CNN achieved the best results, where the Macro F1 score is 51\%  and accuracy is 80\%, as shown in Table 4. It can be noted from the table that training DL models from scratch obtained a good result, but not as good as the results that are recorded in the shared task. DL models alone can not achieve the best results alone, it should be combined with other pre-trained language models to improve the results as you will see in the next experiments.  

% \usepackage{multirow}
\begin{table}[h]
\centering
\caption{DL models results (P= Precision, R= Recall, F1= F1-score, A= Accuracy)}
\begin{tabular}{p{3.6cm}p{0.2cm}p{0.2cm}p{0.2cm}p{0.2cm}p{0.2cm}p{0.2cm}p{0.2cm}p{0.2cm}p{0.2cm}p{0.2cm}}
\toprule
\multirow{2}{*}{Model} &
  \multicolumn{3}{c}{Offensive} &
  \multicolumn{3}{c}{Not offensive} &
  \multicolumn{3}{c}{Macro Avg} &
  \multirow{2}{*}{A} \\ 
  \cline{2-10}
 &
  \multicolumn{1}{c}{P} &
  \multicolumn{1}{c}{R} &
  F1 &
  \multicolumn{1}{c}{P} &
  \multicolumn{1}{c}{R} &
  F1 &
  \multicolumn{1}{c}{P} &
  \multicolumn{1}{c}{R} &
  F1 &
   \\ 
   % \hline
RNN(LSTM) + AraVec &
  \multicolumn{1}{c}{.50} &
  \multicolumn{1}{c}{.05} &
  .10 &
  \multicolumn{1}{c}{.81} &
  \multicolumn{1}{c}{.99} &
  .89 &
  \multicolumn{1}{c}{.65} &
  \multicolumn{1}{c}{.52} &
  .49 &
  .80 \\
  % \hline
CNN + AraVec &
  \multicolumn{1}{c}{.50} &
  \multicolumn{1}{c}{.07} &
  \multicolumn{1}{c}{.13} &
  \multicolumn{1}{c}{.81} &
  \multicolumn{1}{c}{.98} &
  \multicolumn{1}{c}{.89} &
  \multicolumn{1}{c}{.65} &
  \multicolumn{1}{c}{.53} &
  \multicolumn{1}{c}{.51} &
  \multicolumn{1}{c}{.80} \\
  % \hline
LSTM + CNN + AraVec &
  \multicolumn{1}{c}{.38} &
  \multicolumn{1}{c}{.09} &
  .14 &
  \multicolumn{1}{c}{.81} &
  \multicolumn{1}{c}{.96} &
  .88 &
  \multicolumn{1}{c}{.59} &
  \multicolumn{1}{c}{.53} &
  .51 &
  .79 \\ 
\botrule
\end{tabular}
\end{table}

\begin{figure}[htp]
    \centering
    \includegraphics[scale=0.6]{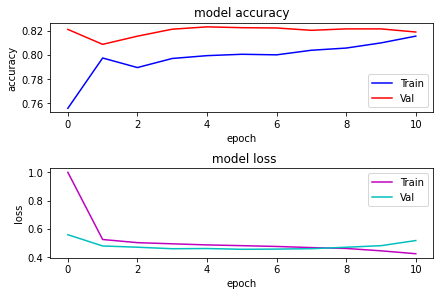}
    \caption{Monitor the performance of training CNN+RNN model}
    \label{fig:perf}
\end{figure}

Generally, these results are very low compared to the rest of the experiments, and the models stop early after a number of epochs as Figure 4 shows, and there is no overfitting. In addition, we notice from Table \ref{fig:perf} that the recall ratio is very low. In our opinion, these Bad results appear due to an imbalance in data that we talked about it in dataset section.
\vspace{5mm} %5mm vertical space
\newpage
\paragraph{Results of transformer models}

In this section, we describe the results for transformer models experiments

% \vspace{5mm} %5mm vertical space

\subparagraph{Results of Arabic Pre-trained language models}

In pre-trained language models experiments, the highest result is from the QARiB model with the AraBert preprocessing experiment, where the macro F1-score is 92\% and accuracy is 95\% as shown in Table 5, and it's the best result we achieved on all experiments. Also in the other experiments with other language models (MARBERT, multi-dialect-bert-base-arabic) we obtained interesting Macro F1 scores and accuracy compared with our other experiments. It can be noted that pre-trained language models improve the classification accuracy. That is one of the advantages of transformer language models that can be adopted for any NLP task.

% \usepackage{multirow}
\begin{table}[h]
\centering
\caption{Transformer model [Language models] results (P= Precision, R= Recall, F1= F1-score, A= Accuracy}

\begin{tabular}{p{3.8cm}p{0.05cm}p{0.05cm}p{0.05cm}p{0.05cm}p{0.05cm}p{0.05cm}p{0.05cm}p{0.05cm}p{0.05cm}p{0.05cm}|}
\toprule
\multicolumn{1}{l}{\multirow{2}{*}{Model}} &
  \multicolumn{3}{c}{Offensive} &
  \multicolumn{3}{c}{Not offensive} &
  \multicolumn{3}{c}{Macro Avg} &
  \multicolumn{1}{c}{\multirow{2}{*}{A}} \\ 
  % \cline{2-10}
  \midrule
\multicolumn{1}{c}{} &
  \multicolumn{1}{c}{P} &
  \multicolumn{1}{c}{R} &
  \multicolumn{1}{c}{F1} &
  \multicolumn{1}{c}{P} &
  \multicolumn{1}{c}{R} &
  \multicolumn{1}{c}{F1} &
  \multicolumn{1}{c}{P} &
  \multicolumn{1}{c}{R} &
  \multicolumn{1}{c}{F1} &
  \multicolumn{1}{c}{} \\ 
  % \hline
QARiB &
  \multicolumn{1}{c}{.88} &
  \multicolumn{1}{c}{.84} &
  \multicolumn{1}{c}{.86} &
  \multicolumn{1}{c}{.96} &
  \multicolumn{1}{c}{.97} &
  \multicolumn{1}{c}{.97} &
  \multicolumn{1}{c}{.92} &
  \multicolumn{1}{c}{.90} &
  \multicolumn{1}{c}{.91} &
  \multicolumn{1}{c}{.94} \\ 
  % \hline
QARiB + AraBert  &
  \multicolumn{1}{c}{.88} &
  \multicolumn{1}{c}{.87} &
  \multicolumn{1}{c}{.87} &
  \multicolumn{1}{c}{.97} &
  \multicolumn{1}{c}{.97} &
  \multicolumn{1}{c}{.97} &
  \multicolumn{1}{c}{.92} &
  \multicolumn{1}{c}{.92} &
  \multicolumn{1}{c}{.92} &
  \multicolumn{1}{c}{.95} \\ 
  % \hline
MARBERT &
  \multicolumn{1}{c}{.88} &
  \multicolumn{1}{c}{.76} &
  \multicolumn{1}{c}{.82} &
  \multicolumn{1}{c}{.94} &
  \multicolumn{1}{c}{.97} &
  \multicolumn{1}{c}{.96} &
  \multicolumn{1}{c}{.91} &
  \multicolumn{1}{c}{.87} &
  \multicolumn{1}{c}{.89} &
  \multicolumn{1}{c}{.93} \\ 
  % \hline
multi-dialect-bert  + AraBert   &
  \multicolumn{1}{c}{.83} &
  \multicolumn{1}{c}{.80} &
  \multicolumn{1}{c}{.81} &
  \multicolumn{1}{c}{.96} &
  \multicolumn{1}{c}{.96} &
  \multicolumn{1}{c}{.96} &
  \multicolumn{1}{c}{.89} &
\multicolumn{1}{c}{.88} &
  \multicolumn{1}{c}{.89} &
  \multicolumn{1}{c}{.93} \\
  % \hline
multi-dialect-bert &
  \multicolumn{1}{c}{.86} &
  \multicolumn{1}{c}{.82} &
  \multicolumn{1}{c}{.84} &
  \multicolumn{1}{c}{.95} &
  \multicolumn{1}{c}{.97} &
  \multicolumn{1}{c}{.96} &
  \multicolumn{1}{c}{.91} &
  \multicolumn{1}{c}{.89} &
  \multicolumn{1}{c}{.90} &
  \multicolumn{1}{c}{.94}\\
  \botrule
\end{tabular}
\end{table}

\vspace{5mm}
\newpage
\subparagraph{Results of hate speech pre-trained models}

In Hate speech Pre-trained models experiments, the best result is from the xlm-r-large-arabic-toxic model with AraBert preprocessing experiment, where the Macro F1 score reached 75\% and the accuracy to 83\% as shown in Table 6. Our experiments with fine-tuning pre-trained models got low results since the best result is 60\% and the accuracy is 67\% from the experiment of a fine-tuning dehatebert-mono-arabic model. It can be noted that the pre-trained model has some limitations to classify new data as it was trained on older data, and even with fine tuning, the performance is still not high, and there is a room for improvement. 

% \usepackage{multirow}

\begin{table}[h]
\centering
\caption{Transformer model [Pre-trained model] results (P= Precision, R= Recall, F1= F1-score, A= Accuracy)}
\label{table:transformer-model}
\begin{tabular}{|l|ccc|ccc|ccc|c|}
\hline
\multirow{2}{*}{Model} & \multicolumn{3}{c|}{Offensive} & \multicolumn{3}{c|}{Not offensive} & \multicolumn{3}{c|}{Macro average} & \multirow{2}{*}{A} \\
 & P & R & F1 & P & R & F1 & P & R & F1 & \\ 
\hline
xlm-r-large-arabic-toxic Fine-tuning & .25 & .45 & .32 & .82 & .66 & .73 & .54 & .55 & .52 & .61 \\
Xlm-r-large-arabic-toxic & .47 & .75 & .57 & .92 & .78 & .85 & .69 & .77 & .71 & .78 \\
xlm-r-large-arabic-toxic AraBert & .56 & .67 & .61 & .91 & .87 & .89 & .74 & .77 & .75 & .83 \\
dehatebert-mono-arabic Fine-tuning & .33 & .64 & .44 & .88 & .67 & .76 & .61 & .66 & .60 & .67 \\
dehatebert-mono-arabic & .34 & .64 & .44 & .88 & .69 & .77 & .61 & .66 & .61 & .68 \\ 
\hline
\end{tabular}
\end{table}

\vspace{5mm}
\newpage
\subparagraph{Results of zero-shot classifier models}

In zero-shot classifier models experiments, the best one is $xlm-roberta-large-xnli-anli$ without any preprocess and with toxic, $not_toxic$ classes, where the Macro F1 score is 68\%, and the accuracy is 80\%, as shown in Table 7.

% \usepackage{multirow}
% \documentclass{article}

\begin{table}[ht]
\centering
\caption{Transformer model [Zero-Shot classifier model (xlm-roberta-large-xnli-anli)] results (P= Precision, R= Recall, F1= F1-score, A= Accuracy)}
\label{table:Zero shot transformer-model}

% \begin{tabularx}{\textwidth}{|X|ccc|ccc|ccc|c|}
% \begin{tabular}{@{}p{5cm}ccc@{}ccc@{}ccc@{}c@{}}
\begin{tabular}{|@{}p{4cm}|ccc|ccc|ccc|c|}

% \begin{tabular}{|l|ccc|ccc|ccc|c|}
\hline
\multirow{2}{*}{Model} & \multicolumn{3}{c}{Offensive} & \multicolumn{3}{c}{Not offensive} & \multicolumn{3}{c}{Macro average} & \multirow{2}{*}{A} \\
 & P & R & F1 & P & R & F1 & P & R & F1 & \\ 

\midrule
Zeroshot, labels: toxic \&
not toxic & .51 & .47 & .48 & .87 & .89 & .88 & .69 & .68 & .68 & .80 \\
Zeroshot + AraBert, labels:
toxic \& not toxic & .54 & .36 & .43 & .85 & .92 & .89 & .69 & .64 & .66 & .81 \\

\bottomrule
\end{tabular}
\end{table}

\begin{table}[ht]
\centering
\caption{Transformer model [Zero-Shot classifier model] results (P= Precision, R= Recall, F1= F1-score, A= Accuracy)}
\label{table:Zero shot transformer-model 2}

\begin{tabular}{|@{}p{4cm}|ccc|ccc|ccc|c|}

% \begin{tabular}{|l|ccc|ccc|ccc|c|}
\hline
\multirow{2}{*}{Model} & \multicolumn{3}{c}{Offensive} & \multicolumn{3}{c}{Not offensive} & \multicolumn{3}{c}{Macro average} & \multirow{2}{*}{A} \\
 & P & R & F1 & P & R & F1 & P & R & F1 & \\ 

\midrule
xlm-roberta-large-xnli +  AraBert, labels:  hate \& Not\_hate  & .54 & .37 & .44 & .85 & .92 & .89 & .70 & .64 & .66 & .81 \\
roberta-large-mnli, labels: toxic \& not\_toxic & .18 & .30 & .22 & .79 & .64 & .71 & .48 & .47 & .47 & .58 \\

\bottomrule
\end{tabular}
\end{table}

In general, comparing macro F1 score and the accuracy results in Tables 4, 5, 6, 7, the QARiB model with AraBERT preprocessor has achieved the best result, where the Macro F1 score is 92\% and the accuracy is 95\%. This result outperformed the best results that are published in the Semeval 2020 shared task, where the best one for the ALAMIHamza team (Alami et al., 2020) obtained 90.17\% in Macro F1-score and 93.9\% in accuracy. Which, by the way, we are using the same dataset splitting that used in the shared task 7000 tweets for training, 1000 tweets for development, and 2000 for testing.

% \newpage
\section{HATE SPEECH MASKING}\label{sec: hatespeech masking}
This section describes the methodology that is used for masking hate speech from content. At the first, we describe the parallel corpus that is built and used in our work. Then, it describes the methodology steps including the model that is used for hate speech masking. Finally, the section presents experimental setups, evaluation metrics, results and discussion.

As mentioned in the introduction, to the best of our knowledge, this problem has not been addressed in the literature, so here we try to open the door for a new research direction to address this new task.

\begin{figure}[htp]
    \centering
    \includegraphics[scale=0.3]{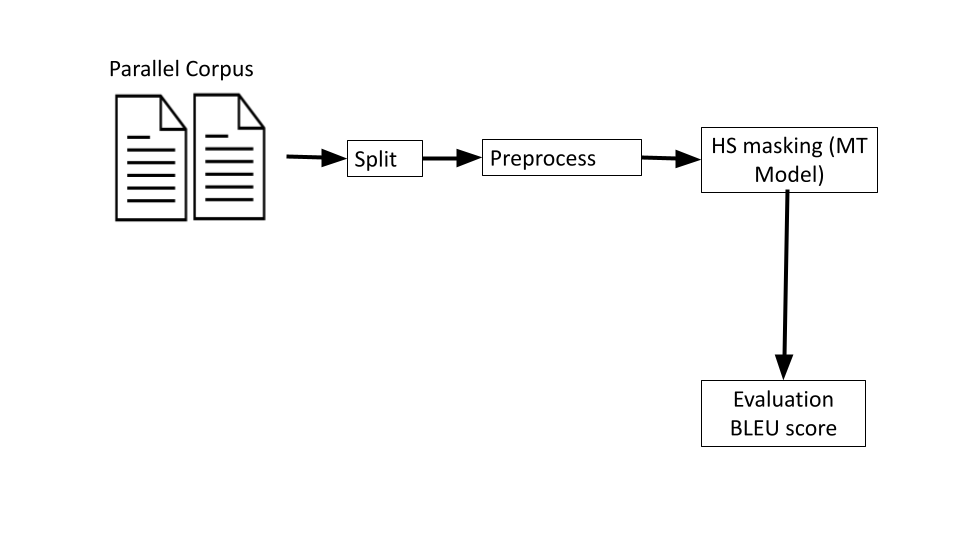}
    \caption{The Brief of Hate Speech Masking Methodology}
    \label{fig:hs_masking}
\end{figure}

Figure \ref{fig:hs_masking} presents a brief overview of the methodology phases including parallel corpus preparation, data preprocessing, training model, and evaluation metrics.

\subsection{Parallel corpus preparing}
\begin{figure}[htp]
    \centering
    \includegraphics[scale=1.2]{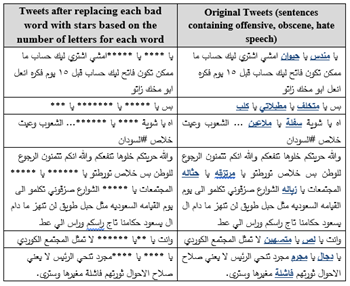}
    \caption{Parallel corpus samples}
    \label{fig:parallel}
\end{figure}

We build a parallel corpus that contains pairs of sentences as shown in Figure \ref{fig:hs_masking}, the first part containing part from the dataset that is published in the shared task SemEval-2020 \cite{Zampieri} for Arabic offensive language detection, the second part contains the same sentences in the first pair with masking the bad words in the sentences, by replacing each bad word with stars based on the number of letters for each word, which requires human workers to mask the dirty word.
In this step, a volunteer and prepared the second part from the parallel corpus, where she masked bad words from sentences for the entire dataset as shown in Figure \ref{fig:parallel}.

\subsection{Split parallel corpus}
We have adopted two different sizes of data in our strategy, the first one contains 3183 pairs partitioned into 1992 pairs classified as hate speech and 1592 pairs classified as not hate speech. And the second one contains 4783 pairs partitioned into 1992 pairs classified as hate speech and 2791 pairs classified as not hate speech. Each group of datasets is divided for training set, development set, and testing set as shown in Table 8

\begin{table}[h]
\caption{Parallel corpus groups Details}
% \usepackage{booktabs}

% \begin{tabular}{@{}|l|l|l|l|l|@{}}
% \toprule
\begin{tabular}{p{2.8cm}p{0.05cm}p{0.05cm}p{0.05cm}p{0.05cm}}
\toprule

             & \multicolumn{1}{c}{Training set} &
             \multicolumn{1}{c}{Development set} & 
             \multicolumn{1}{c}{Test set} &\multicolumn{1}{c}{ Total} \\ \midrule
\multicolumn{1}{c}{First group}  & \multicolumn{1}{c}{2388}         & \multicolumn{1}{c}{795}             & \multicolumn{1}{c}{401}      & \multicolumn{1}{c}{3183}  \\ 
\multicolumn{1}{c}{Second group} & \multicolumn{1}{c}{3287}         & \multicolumn{1}{c}{1095}            & \multicolumn{1}{c}{401}      & \multicolumn{1}{c}{4783}  \\ \bottomrule
\end{tabular}
\end{table}

\subsection{Dataset preprocessing}
We apply same preprocessing steps described in section 3.2.
\subsection{Hate speech masking model}
We build a hate speech masking model using a neural machine translation with a transformer model since we consider the problem as a machine translation problem. 

The model starts with parsing the data, each line contains a sentence that contains bad words and its corresponding same sentence that masking to bad words. The sentence that contains bad words is the source sequence and the same sentence with the bad word mask is the target sequence. We prepend the token "[start]" and we append the token "[end]" to the target sentence. Then, the model uses two instances of the TextVectorization layer to vectorize the text data, which it to turn the original strings into integer sequences where each integer represents the index of a word in a vocabulary. Then, building the architecture of the sequence-to-sequence transformer which consists of a Transformer Encoder and a Transformer Decoder chained together. The source sequence will be passed to the transformer encoder, which will produce a new representation of it. This new representation will then be passed to the transformer decoder then will seek to predict the next words in the target sequence.  

The Transformer Decoder receives the entire sequences at once, and thus we must make sure that it only uses information from target tokens 0 to N when predicting token N+1. After that, we training model, we used 64 batch size, embedding dimension 256, and 30 epochs. We used early stopping on the validation set with patience 2 to terminate training when the validation loss has stopped decreasing after two epochs with no improvement. And we add L2 regularization which is a technique to reduce the complexity of the model. It does so by adding a penalty term to the loss function. 

\subsection{Evaluation Metrics}
In this section, we describe the evaluation metric that is used in our work. Since the methodology that we use considers the problem as a machine translation problem, we use BLEU Score for evaluation, which is defined as an algorithm for evaluating the quality of text which has been machine-translated from one natural language to another \cite{Papineni}. The primary assumption behind BLEU is that the closer a machine translation is to a professional human translation, the better it is. BLEU was one of the first metrics to claim a high correlation with human judgments of quality \cite{coughlin-2003-correlating} and remains one of the most popular automated and inexpensive metrics. In our experiments, we use BLEU to compare the generated text with the reference test text sets. The BLEU score calculations allow you to specify the weighting of different n-grams in the calculation of the BLEU score. This gives you the flexibility to calculate different types of BLEU scores, such as individual and cumulative n-gram scores.
In our work, we use the same BLEU score implementation that is available by \cite{SharmaASZ17} to evaluate our experiments with two data set sizes and with various vocabulary sizes.

\subsection{Experiments and Results}
We executed a set of experiments to evaluate the proposed models for Arabic hate speech Masking using a machine translation model. 

In the dataset that has 3183 sentences as explained in Table 8, which contain 1592 from it are classified as not hate sentences, the best BLEU score we got is (0.29, 0.17, 0.10, 0.07) for (1-gram, bi-gram, 3-gram, 4-gram) sequentially, when the vocabulary size is 8000 words, as shown in Table 9. The Not-HS size column displays the number of sentences that are classified as not hate from each dataset, and the UNK column represents the number of unknown words for each experiment after prediction.

% \usepackage{multirow}
\begin{table}[h]
\caption{BLEU score for five experiments trained on dataset size equal 3183 with various vocabulary sizes (DS = Dataset, Not-HS = Not hate speech size, UNK = unknown,BLEU 1 = BLEU 1-Gram,BLEU 2 = BLEU 2-Gram,BLEU 3 = BLEU 3-Gram,BLEU 4 = BLEU 4-Gram)}
% \usepackage{multirow}
% \usepackage{multirow}
\begin{tabular}{p{0.05cm}p{0.05cm}p{0.05cm}p{0.05cm}p{0.05cm}p{0.05cm}p{0.05cm}p{0.05cm}}
\toprule
% {|pp{0.05cm}|p{1.5cm}|p{1.5cm}|p{1.5cm}|pp{0.05cm}|pp{0.05cm}|pp{0.05cm}|pp{0.05cm}|}

\multicolumn{1}{c}{DS size}           &
\multicolumn{1}{c}{Not-HS size}& 
\multicolumn{1}{c}{vocab\_size} & 
\multicolumn{1}{c}{UNK} &
\multicolumn{1}{c}{BLEU 1} &
\multicolumn{1}{c}{BLEU 2} &
\multicolumn{1}{c}{BLEU 3} & 
\multicolumn{1}{c}{BLEU 4} \\ \hline
                      \multirow{5}{*}{3183} & 
                      \multirow{5}{*}{1592}                 & \multicolumn{1}{c}{15000}      & \multicolumn{1}{c}{1537}                                                        & \multicolumn{1}{c}{0.25}        & \multicolumn{1}{c}{0.14}        & \multicolumn{1}{c}{0.08}        & \multicolumn{1}{c}{0.05}        \\  
                       &                                                                                    & \multicolumn{1}{c}{12000}      & \multicolumn{1}{c}{1957}                                                          & \multicolumn{1}{c}{0.22}        & \multicolumn{1}{c}{0.11}        & \multicolumn{1}{c}{0.06}        & \multicolumn{1}{c}{0.03}        \\  
                       &                                                                                    & \multicolumn{1}{c}{10000}      & \multicolumn{1}{c}{2853}                                                          & \multicolumn{1}{c}{0.27}        & \multicolumn{1}{c}{0.14}        & \multicolumn{1}{c}{0.08}        & \multicolumn{1}{c}{0.05}        \\  
                       &                                                                                    & \multicolumn{1}{c}{8000}       & \multicolumn{1}{c}{1707}                                                          & \multicolumn{1}{c}{0.29}        & \multicolumn{1}{c}{0.17}        & \multicolumn{1}{c}{0.10}        & \multicolumn{1}{c}{0.07}        \\  
&                                                           & \multicolumn{1}{c}{6000}       &
\multicolumn{1}{c}{3246}                                                          &
\multicolumn{1}{c}{0.28}        & 
\multicolumn{1}{c}{0.16}        & 
\multicolumn{1}{c}{0.10}        & 
\multicolumn{1}{c}{0.06}        \\ \hline
\end{tabular}
\end{table}

In the dataset that has 4383 sentences as explained in Table 8, which contain 2791 from it are classified as not hate sentences, the best BLEU score we got is (0.30, 0.18, 0.12, 0.08) for (1-gram, bigram, 3-gram, 4-gram) sequentially, when the vocabulary size is 12000 words, as shown in Table 10.

% \usepackage{booktabs}
% \usepackage{multirow}

\begin{table}[h]
\caption{BLEU score for five experiments trained on dataset size equal 4383 with various vocabulary sizes (DS = Dataset, Not-HS = Not hate speech size, UNK = unknown,BLEU 1 = BLEU 1-Gram,BLEU 2 = BLEU 2-Gram,BLEU 3 = BLEU 3-Gram,BLEU 4 = BLEU 4-Gram).}
\begin{tabular}{p{0.05cm}p{0.05cm}p{0.05cm}p{0.05cm}p{0.05cm}p{0.05cm}p{0.05cm}p{0.05cm}}

\toprule

\multicolumn{1}{c}{DS size}           &
\multicolumn{1}{c}{Not-HS size}& 
\multicolumn{1}{c}{vocab\_size} & 
\multicolumn{1}{c}{UNK} &
\multicolumn{1}{c}{BLEU 1} &
\multicolumn{1}{c}{BLEU 2} &
\multicolumn{1}{c}{BLEU 3} & 
\multicolumn{1}{c}{BLEU 4} \\ \midrule
                      \multirow{5}{*}{4383} & 
                      \multirow{5}{*}{2791}       & 
                      \multicolumn{1}{c}{15000}      & 
                      \multicolumn{1}{c}{2712} & 
                      \multicolumn{1}{c}{0.28}        &
                      \multicolumn{1}{c}{0.16}        & 
                      \multicolumn{1}{c}{0.10}        & \multicolumn{1}{c}{0.06}        \\ 
                       &                        & \multicolumn{1}{c}{12000}      & \multicolumn{1}{c}{1542} & \multicolumn{1}{c}{0.30}        & \multicolumn{1}{c}{0.18}        & \multicolumn{1}{c}{0.12}        & \multicolumn{1}{c}{0.08}        \\ 
                       &                        & \multicolumn{1}{c}{10000}      & \multicolumn{1}{c}{2027} & \multicolumn{1}{c}{0.26}        & \multicolumn{1}{c}{0.16}        & \multicolumn{1}{c}{0.10}        & \multicolumn{1}{c}{0.06}        \\ 
                       &                        & \multicolumn{1}{c}{8000}       & \multicolumn{1}{c}{2402} & \multicolumn{1}{c}{0.29}        & \multicolumn{1}{c}{0.17}        & \multicolumn{1}{c}{0.11}        & \multicolumn{1}{c}{0.07}        \\ 
& & \multicolumn{1}{c}{6000}       & \multicolumn{1}{c}{2764} & \multicolumn{1}{c}{0.29}        & \multicolumn{1}{c}{0.18}        & \multicolumn{1}{c}{0.11}        & \multicolumn{1}{c}{0.07}        \\ \bottomrule
\end{tabular}
\end{table}

In general, comparing the BLEU score in Table 9,10, the model with dataset size 4383 and vocabulary size 12000 has achieved the best result, where BLEU score with 1-gram is 30\%, which is a good result compared with the state of the art MT systems. It can be noted from the Table that this task is very challenging, but we open a new research direction with this new task.
% \newpage
\section{CONCLUSION}\label{sec: conclusion}

In this paper we handle two problems: the first is Arabic hate speech detection using different neural networks architectures including RNN, CNN, and Transformers, and the second problem of cleaning offensive/hate speech texts. The dataset used is from the shared task SemEval-2020 \cite{Zampieri} for Arabic offensive language detection. In the hate speech detection task, we conduct several experiments to find the best model by checking the best macro F1 score and accuracy. The best Macro F1 score with 92\% and accuracy of 95\% was obtained by the QARiB model with the AraBERT preprocessor. And in cleaning hate speech texts, we use BLEU Score for evaluation, based on considering the problem of cleaning dirty text as a machine translation problem, and the best result achieved is 30\% with 1-gram, which is achieved with dataset size 4383 and vocabulary size 12000 as explained in section 5.2. As a summary of our work, the result of one of our experiments in hate speech detection outperformed the best results that are published in the Semeval 2020 shared task.  And to the best of our knowledge, we worked on the first experiment in Arabic hate speech masking as a machine translation model, and it achieved a good result compared with the state of the art  MT systems.

For future work, the parallel corpus that we use in the hate speech masking model will be increased because that will increase the BLEU score as we noted in our experiments, build web applications to deploy the hate speech detection and masking models, and publish the lexicon that we extracted when we built the parallel corpus, and which include hate/offensive words with the category for each of them. In additional, build a model for hate speech paraphrasing using a machine translation model.

\section*{Declarations}

\paragraph{Funding}

The authors declare that they have no funding.

\paragraph{Competing Interests}

The authors declare that they have no competing interests.

\paragraph{Authors contribution statement}

Salam conducted the literature review, implemented the proposed models, performed the experiments, and wrote the manuscript. Motaz provided overall project supervision, contributed to the selection of models, and was involved in manuscript editing and linguistic revisions. Furthermore, Salam and Motaz collaboratively analyzed and interpreted the experimental results. All contributing researchers have reviewed and approved the manuscript for submission.

\paragraph{Ethics approval}

This study did not involve data collection from participants nor did it rely on data that requires special ethical approval. All data used were publicly available or obtained through accredited sources that comply with ethical standards for scientific research. 

\paragraph{Availability of data and materials}

All related materials for this paper are available in the GitHub repository at \hyperlink{https://github.com/motazsaad/hate-speech}{https://github.com/motazsaad/hate-speech}. Additional details can be provided by the corresponding author upon reasonable request.

% \section*{References}

\bibliography{sn-article}

\end{document}